\documentclass[a4paper]{article}

\usepackage{latexsym}
\usepackage{amsmath}
\usepackage{amssymb}
\usepackage{amsfonts}
\usepackage{dialogue}
\usepackage{graphicx}
\usepackage{subcaption}
\graphicspath{ {./images/} }

\usepackage{INTERSPEECH2021}

\title{HarperValleyBank: A Domain-Specific Spoken Dialog Corpus}
\name{Mike Wu$^1$, Jonathan Nafziger$^2$, Anthony Scodary$^2$, Andrew Maas$^1$}
\address{
  $^1$Stanford University\\
  $^2$Gridspace, Inc.}
\email{\{wumike,amaas\}@cs.stanford.edu, \{jonathan,anthony\}@gridspace.com}

\begin{document}

\maketitle
\begin{abstract}
We introduce \textsc{HarperValleyBank}, a free, public domain spoken dialog corpus. The data simulate simple consumer banking interactions, containing about 23 hours of audio from
1,446 human-human conversations between 59 unique speakers. We selected intents and utterance templates to allow realistic variation while controlling overall task complexity and limiting vocabulary size to about 700 unique words. We provide audio data along with
transcripts and annotations for speaker identity, caller intent, dialog actions, and
emotional valence. The data size and domain specificity makes for
quick transcription experiments with modern end-to-end neural approaches.
Further, we provide baselines for
representation learning, adapting
recent work to embed waveforms for downstream prediction tasks.
Our experiments show that tasks using our annotations are sensitive to both the model choice and corpus size.
 % for representation learning approaches.
\end{abstract}
% \noindent\textbf{Index Terms}: speech recognition, human-computer interaction, computational paralinguistics

\section{Introduction}
Recent innovations in deep learning approaches substantially improved spoken dialog
systems in both academic research and industry applications. Speech recognition
systems now regularly leverage neural network models to achieve near human performance \cite{hinton2012deep, saon2017english, xiong2018microsoft, XXXciteConformer}.
Modern systems increasingly use architecture themes including attention and end-to-end recurrent neural networks to encode few assumptions and rapidly adapt to new data \cite{hannun2014deep,
chan2016listen, maas2015lexicon, likhomanenko2019needs, graves2014towards, bahdanau2016end}. In parallel, approaches to spoken and text-based dialog
systems increasingly leverage neural networks for dialog management and
state representation \cite{khatri2018advancing, andreas2016learning, liu2017iterative}.

We developed the
\textsc{HarperValleyBank} corpus for homeworks and projects in
Stanford's \emph{Spoken Language Processing} course\footnote{\tt{cs224s.stanford.edu}}, as well as for research on
speech recognition combined with other spoken language tasks. The goals of the corpus are to provide:
\begin{itemize}
    \item Freely available data for education, research, or commercial development. We release the data using a Creative
Commons license (CC-BY)\footnote{\tt{creativecommons.org/licenses/by/4.0}}.

    \item Sufficient size and variability to meaningfully evaluate end-to-end neural approaches for speech transcription.

    \item Manageable overall size and complexity to enable students to quickly iterate on experiments without requiring expensive compute hardware for training.

    \item Annotations for dialog-relevant tasks aside from speech transcription (e.g. intent, dialog action) to enable multi-task training and representation transfer.

    \item Realistic domain-specific, goal-oriented conversations to evaluate representation transfer approaches across domains in spoken dialog systems.
\end{itemize}

% Building and experimenting with neural network approaches often requires sufficiently large
% datasets and significant computational resources for training and evaluation.
There is significant recent work in representation learning for domain and task
transfer via embedding models \cite{oord2018representation,chorowski2019unsupervised,schneider2019wav2vec,baevski2020wav2vec}, which can be trained without any supervision and reused for many downstream tasks like predicting speaker identity \cite{panayotov2015librispeech,nagrani2017voxceleb} and commands \cite{warden2018speech,lugosch2019speech}. Representation transfer is important to warm start deep learning based dialog systems on new task domains.
Part of our motivation in designing this corpus is providing a realistic test case for representation learning approaches to adapt to speech tasks. See \cite{serban2015survey} for a review of availaable dialog datasets. We recorded two-sided phone conversations to simulate customer call center interactions in a financial services domain. The dataset is representative of human to human
goal-oriented dialogs for consumer banking with a narrowly scoped set of intents.

In the next sections, we provide more details on the corpus and its collection, followed by experiments showcasing its applications to automatic speech recognition and unsupervised representation learning.
In Sec.~\ref{sec:corpus}, we discuss basic corpus statistics, caller intents, and the data generation and annotation process.
In Sec.~\ref{sec:asr}, we explore end-to-end neural models with multi-task objective functions to simultaneously perform speech-to-text transcription and caller intent prediction.
In Sec.~\ref{sec:transfer}, we explore using caller intents and sentiments as downstream objectives to evaluate representation transfer, and report  unsupervised baselines. The full dataset with a PyTorch implementation reproducing the
speech recognition and transfer experiments is publically available\footnote{https://github.com/cricketclub/gridspace-stanford-harper-valley}.

\section{The \textsc{HarperValleyBank} Corpus}
\label{sec:corpus}

We compile a dataset of recorded audio conversations between an agent and a
customer of a bank. Conversations are goal-oriented, such as ordering a
new checkbook or checking the balance of an account. Fig.~\ref{diag:ex1}
shows an example conversation from the dataset. We collected data using the
Gridspace Mixer platform, where crowd workers are randomly paired for short
telephone conversations.
Mixer membership includes hundreds of past and current professional call center
agents who are trained to perform assorted Mixer tasks in domains including
healthcare, telecommunications, and commerce.

\subsection{Data Collection Procedure}
 Using the Mixer web platform, a
person is randomly assigned the role of agent or customer and provided a script
for the interaction along with a telephone number to call to start the
conversation. Roles are randomly assigned for each call, so the same worker can
appear as both customer and agent in different conversations. We created a set
of conversation goals and scripts for each interaction using templates intended
to capture variety in each intent while keeping workers' word choices and the
overall interactions fairly simple with limited vocabulary. We do not control
the noise environment or microphones used by each worker, and there is
natural variation across different types of phones and environments.

When a person calls in, they are paired with the
next available conversation partner. The groups are large enough that many
unique pairings occur over the course of one session.
Once the Mixer task is live for the caller, the web application will change
state, informing the caller whether they are acting in the role of the agent or caller.
The instructions, data, and user interface adapt to the role and provide a rough script.
The customer role initiates a call task by expressing an intent, and the agent
role has an interactive web interface to simulate completing a task.
We encouraged callers to use mobile phones or headsets to encourage a microphone
transfer function that is acoustically representative of a real call center.

During each
call, participants have the script for their side of the conversation in front of
them in a web browser.
% We provide examples of suggested phrasings in the Supplemental Material.
The worker playing the customer role is given a single intent for the conversation, along with specific values for
relevant slots (e.g. the amount of money to transfer and the source/target
accounts). When playing the agent role, a worker is shown some
simple buttons and menus they must click to perform the requested operation
(e.g. "check account balance").

A conversation is deemed successful and considered for the dataset if the
agent correctly executes the task provided to the customer caller. Names and
slot values for different transactions are randomly generated, and we limit the
number of possible names and proper nouns to reduce overall corpus vocabulary
size. The Gridspace Mixer platform handles generating random templates from a
high level specification, all associated telephony operations to pair
callers, and recording audio along with metadata for each interaction.

%\subsection{Data Generation}
%The caller is then assigned \textit{one} of the following of the 8 conversation tasks.
% \begin{itemize}
%   \item Replace Card
%   \item Transfer Money
%   \item Check Balance
%   \item Order Checks
%   \item Pay Bill
%   \item Reset Password
%   \item Schedule Appointment
%   \item Get Branch Hours
% \end{itemize}
%To encourage limited vocabulary, the callers were given a suggested phrasing.
 % This is ordinarily the opposite of what Gridspace instructs it workers to do, so it required them to adjust.
%We provide examples of suggested phrasings in the Supplemental Material.

%
%\begin{table}[h!]
%\centering
%\begin{tabular}{ll}
%    \hline
%  Task & Example Suggested Phrasing \\
%    \hline
%  Greeting & ``Hello, this is Harper Valley \\
%             & National Bank. My name is \\
%             & Mary.'' \\
%  Replace Card & ``Which card would you like \\
%                 & to replace?'' \\
%  Reset Password & ``What is your phone \\
%                  & number?''\\
%    \hline
%\end{tabular}
%\caption{Suggested phrases for conversation intents.  See the supplemental
%  material for a full list of the phrases suggested to Mixers.}
%\end{table}
%
%Once the agent has submitted the task in her user interface and the caller has indicated they have no further actions, the call ends. Both annotators' user interfaces return to the initial state, allowing them to initiate another call if they choose to.

\begin{figure}[tbh]
    \begin{dialogue}
        \speak{agent} hello this is harper valley national bank my name is jay how can I help you today
        \speak{caller} hi my name is mary davis
        \speak{caller} [noise]
        \speak{caller} i would like to schedule an appointment
        \speak{agent} yeah sure what day what time
        \speak{caller} thursday one thirty p m
        \speak{agent} that's done anything else
        \speak{caller} that's it
        \speak{agent} have a good one.
    \end{dialogue}
    \caption{Example conversation from corpus.}
    \label{diag:ex1}
\end{figure}
%\begin{figure}[h!]
%    \begin{dialogue}
%        \speak{agent} hello this is uh harper valley national bank my name is james
%        \speak{caller} hi my name is john johnson
%        \speak{caller} i would like to pay a bill
%        \speak{agent} uhm [unk] yes of course uh what is the company name
%        \speak{caller} the company name is fossil gas
%        \speak{agent} okay and what is the company address
%        \speak{caller} the address is one fifty three main street
%        \speak{caller} harper valley
%        \speak{caller} oregon
%        \speak{caller} five zero nine zero zero
%        \speak{agent} and what is the bill amount
%        \speak{caller} the amount of the bill is ninety one dollars
%        \speak{agent} Okay is there anything else i can help you with
%        \speak{caller} [noise] no
%    \end{dialogue}
%    \caption{Example Conversation}
%    \label{diag:ex2}
%\end{figure}

% \begin{figure}[h!]
%     \begin{dialogue}
%         \speak{agent} hello this is harper valley national bank my name is patricia how can i help you today
%         \speak{caller} hi my name is elizabeth johnson i lost my credit card can you send me a new one
%         \speak{agent} okay which card would you like to replace
%         my credit card
%         \speak{agent} i've ordered your replacement credit card
%         \speak{agent} is there anything else i can help you with
%         \speak{caller} no thank you
%         \speak{agent} thank you for calling have a great day
%     \end{dialogue}
%     \caption{Example Conversation}
%     \label{diag:ex3}
% \end{figure}

\subsection{Data Labelling}

Gridspace Mixer trains a subset of its community to perform a wide
range of annotations. For this corpus, Mixers performed
three primary labeling tasks: \textit{text transcript, audio quality, and
script adherence ratings}.
Gridspace has provided the Mixer community with a highly specialized speech
labeling tool called Scriber. Scriber is designed for rapid human transcription
and data labeling.
The tool also provides a wide array of convenience and ergonomic functions,
designed to enable efficient labeling of large spoken language datasets. Every
person trained to use Scriber must go through several training sessions, which
requires them to watch training videos and perform well on a quiz. For dialog
actions and emotional valence, labels were instead produced using a Gridspace API
rather than human annotation. As a result, there may be some noise or bias, but
our experiments indicate they are reasonable for a benchmark task.

The \textsc{HarperValleyBank} corpus was collected over three separate Mixer
sessions and then filtered post-annotation, informed by
the script adherence labels and audio quality labels, to ensure the data was
simple and low variance. This filtering ensures the corpus provides
conversational and task-oriented speech data while
regulating for simplicity.
The primary target of the cleaning were conversations where calls dropped or
there were other technical issues which derailed the conversation.
Specifically we removed conversations with script adherence ratings less than 4 and audio
quality ratings less than 3.  Furthermore we filtered out conversations which
contained some words such as 'frozen', 'website', and 'refresh', which
indicated conversation about technical issues with the task interface.
In total we removed 375 conversations.

\subsection{Corpus Statistics}

\begin{table}[tbh]
\centering
\begin{tabular}{l l}
\hline
\textsc{HarperValleyBank} Statistics & \\
\hline
Hours of audio & 23.7 \\
\# of conversations/transcripts & 1,446 \\
\# of utterances & 25,730 \\
\# of unique words & 735 \\
Mean \#  of lines per conversation & 17.8 \\
Median \#  of lines per conversation & 16 \\
Mean \#  of words per utterance & 4.1 \\
Median \#  of words per utterance & 4.5 \\
\# of unique speakers & 59 \\
\# of task classes & 8 \\
\# of dialog action classes & 16 \\
\# of sentiment classes & 3 \\
\hline
\end{tabular}
\caption{Basic corpus statistics .}
\label{table:stat}
\end{table}

\begin{figure}[tb]
    \centering
    \begin{subfigure}[b]{0.49\linewidth}
        \centering
        \includegraphics[width=\textwidth]{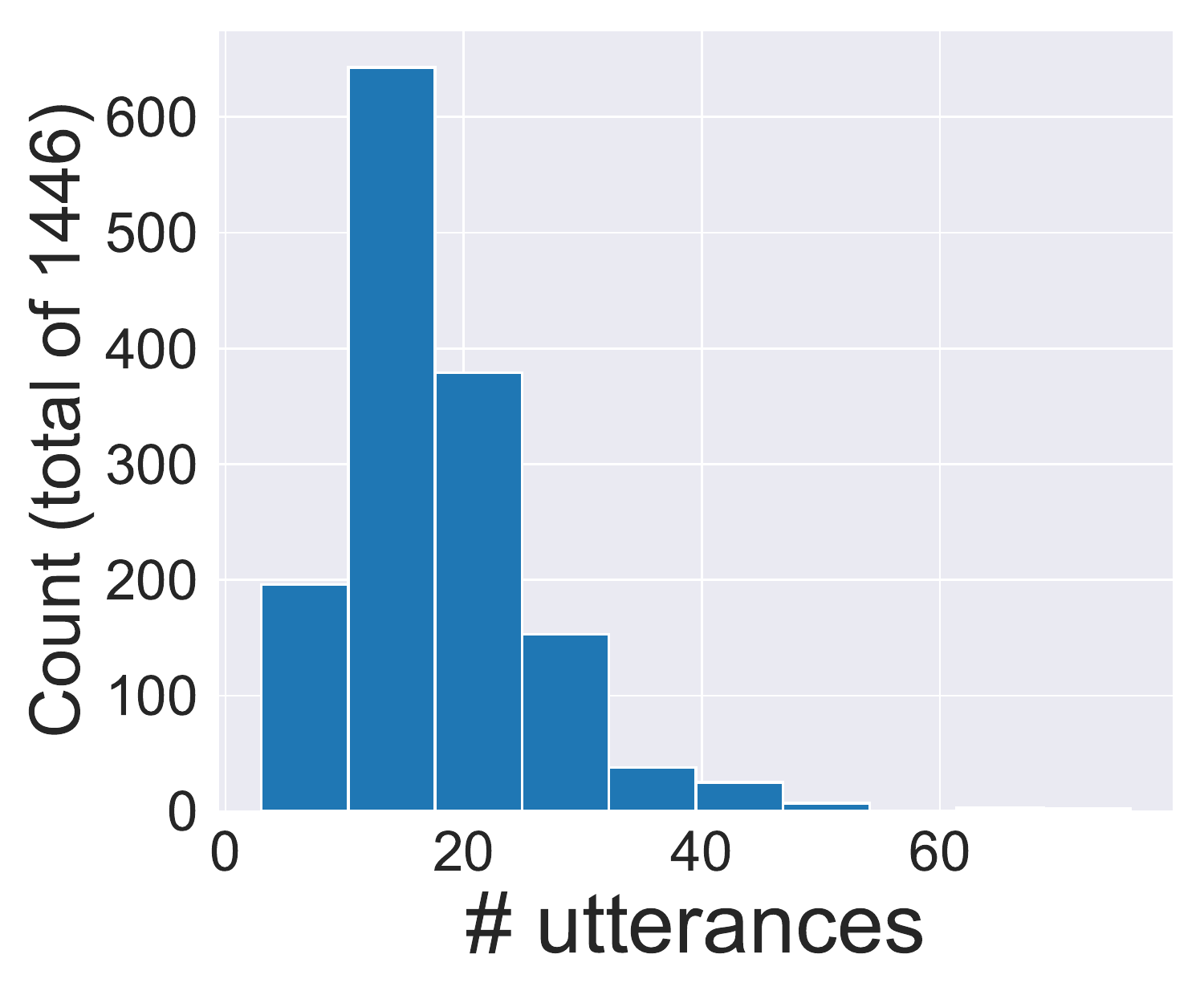}
        \caption{}
    \end{subfigure}
    \begin{subfigure}[b]{0.49\linewidth}
        \centering
        \includegraphics[width=\textwidth]{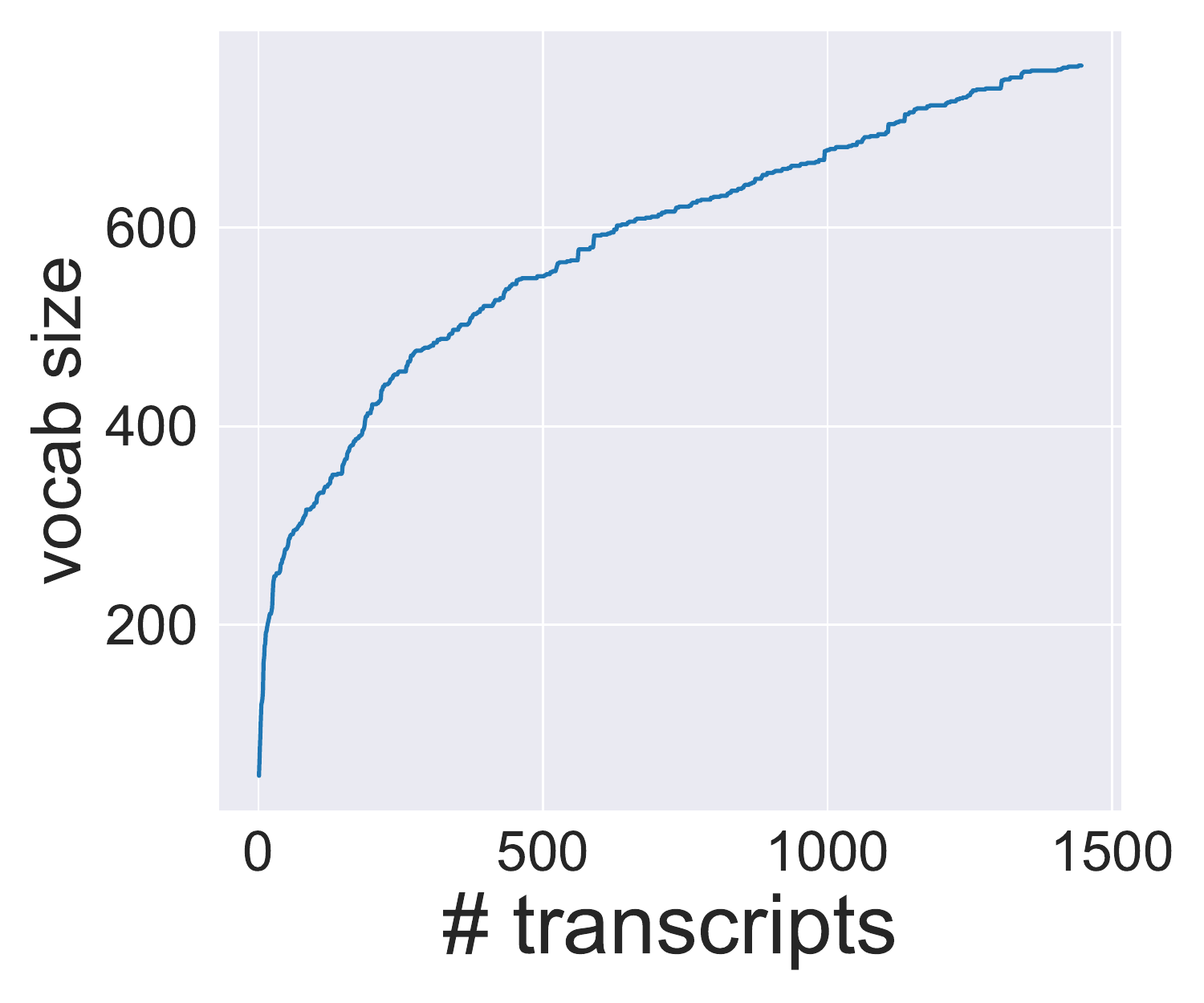}
        \caption{}
    \end{subfigure}
    \caption{(a) The empirical distribution over the number of utterances in a conversation transcript. (b) The number of unique words spoken as a function of the conversation count.}
    \label{fig:stats}
\end{figure}

Table~\ref{table:stat} shows basic statistics of the \textsc{HarperValleyBank} corpus. The corpus contains about 23
hours of audio in total, across 1,446 conversations.
Conversations range from 2 to 60 utterances, with an average of 18. Each
utterance roughly corresponds to a single turn in the conversation. Due to automatic segmentation of utterances, there can be multiple utterances
in a row from a single speaker's turn.
Notably, the corpus has a small vocabulary of approximately 700 unique words.
Many of the most common words in the vocabulary are domain-specific to customer
service e.g. ``help'', ``thank'', or ``please''.
Fig.~\ref{fig:stats}b depicts how vocabulary size scales with dataset size.

\begin{figure}[tb]
    \centering
    \begin{subfigure}[b]{0.50\linewidth}
        \centering
        \includegraphics[width=\textwidth]{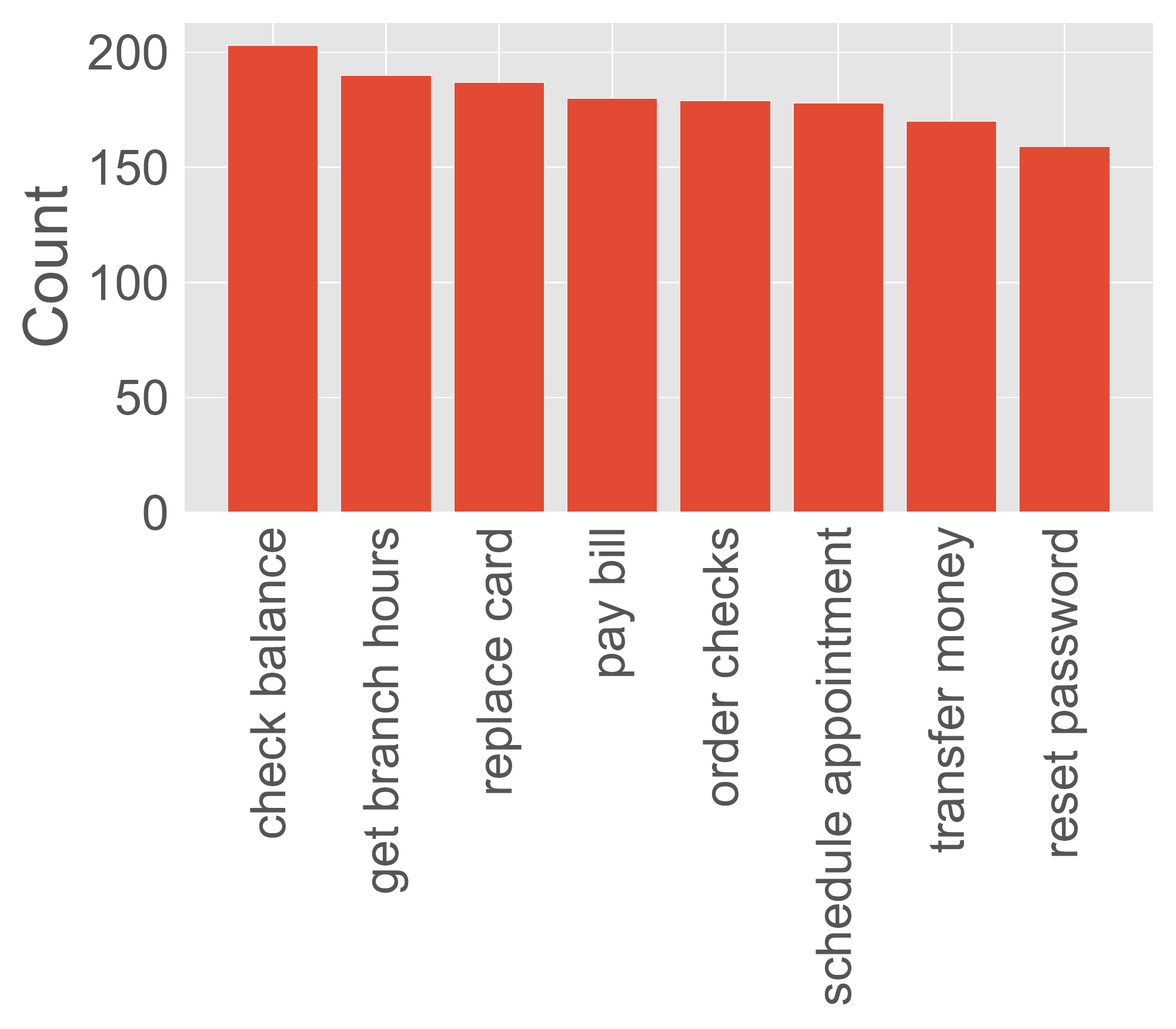}
        \caption{Intent}
    \end{subfigure}
    \begin{subfigure}[b]{0.44\linewidth}
        \centering
        \includegraphics[width=\textwidth]{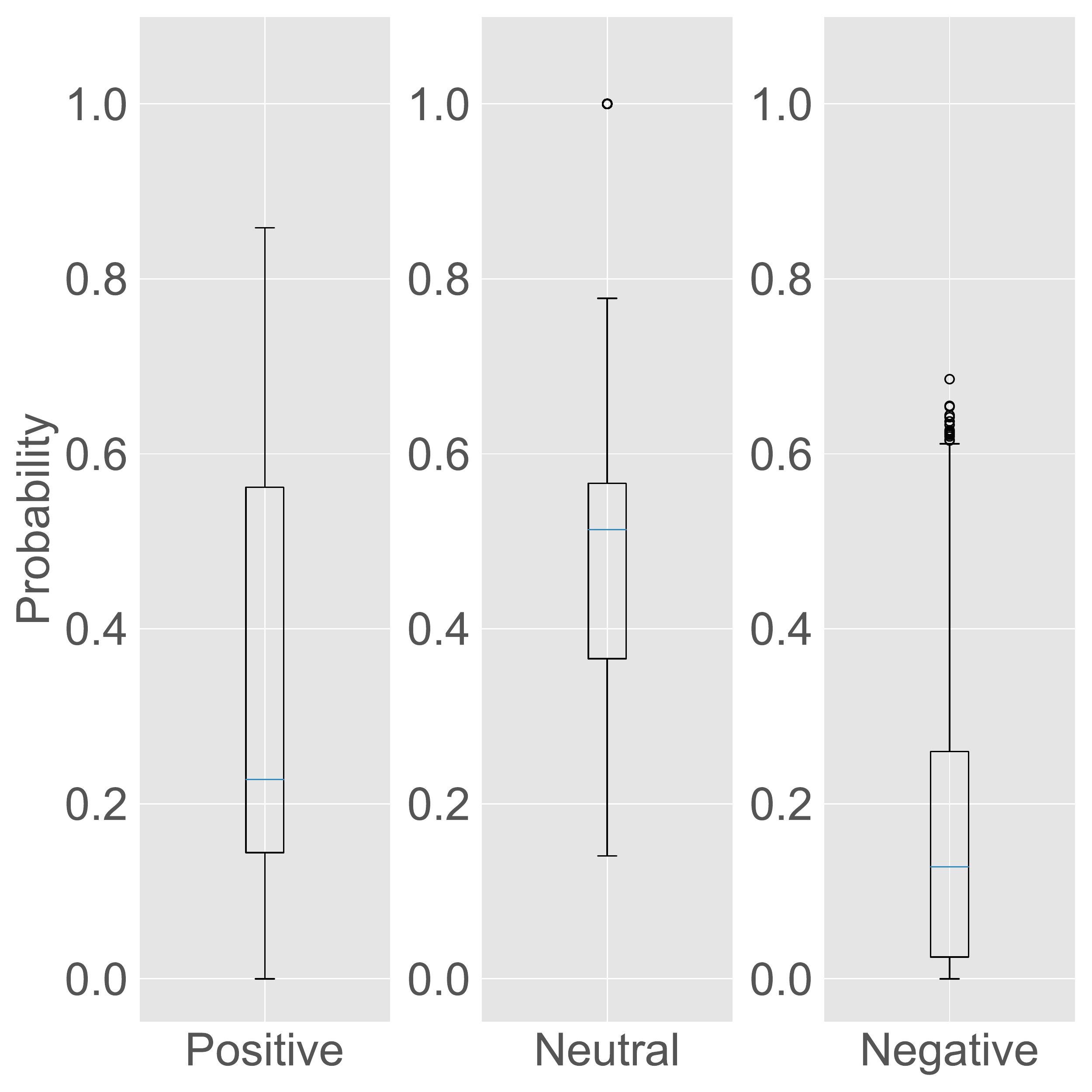}
        \caption{Sentiment}
    \end{subfigure}
    % \begin{subfigure}[b]{\linewidth}
    %     \centering
    %     \includegraphics[width=\textwidth]{speaker_dist.pdf}
    %     \caption{Speaker Identity}
    % \end{subfigure}
    \begin{subfigure}[b]{\linewidth}
        \centering
        \includegraphics[width=\textwidth]{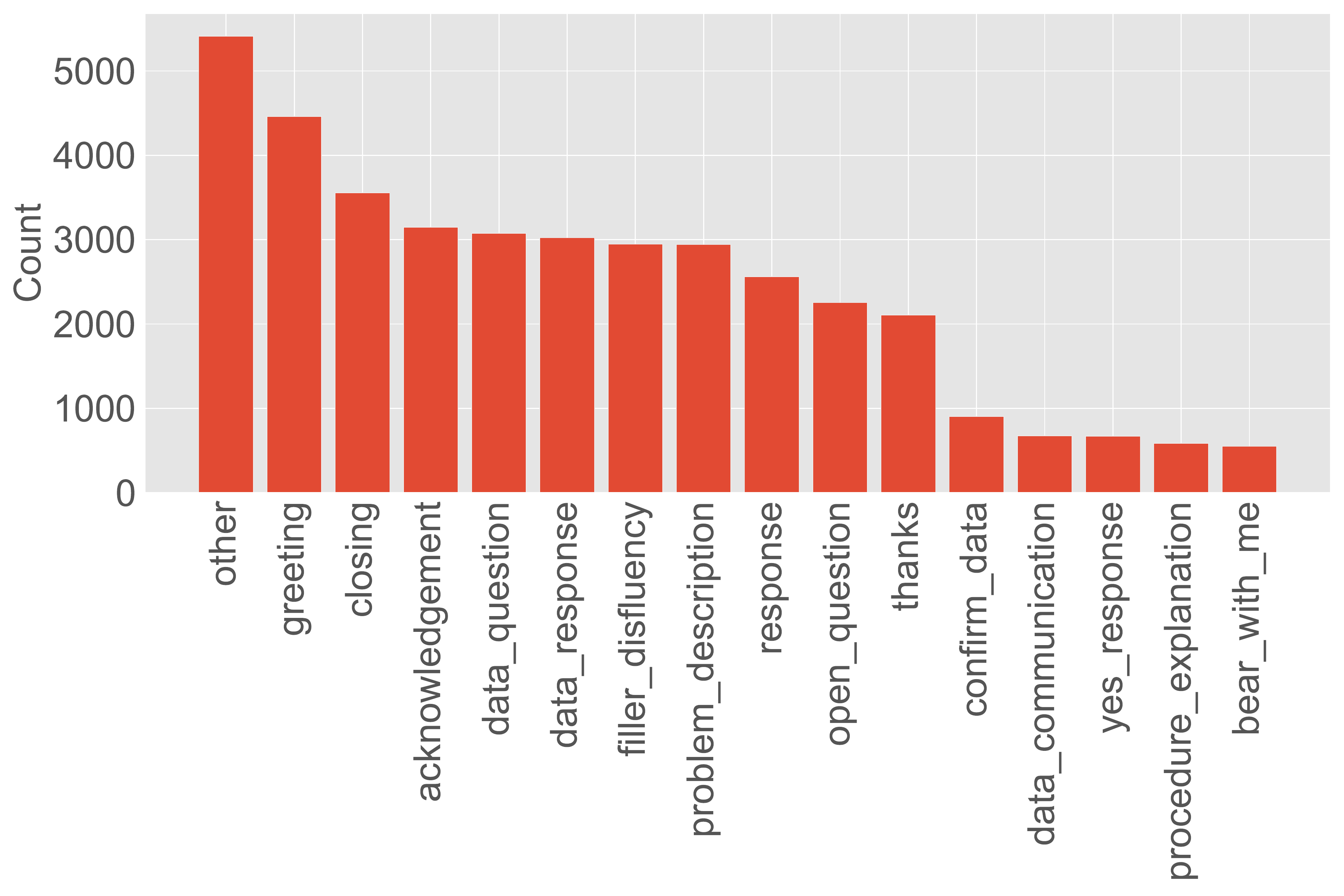}
        \caption{Dialog Actions}
    \end{subfigure}
    \caption{Distribution over auxiliary labels. Subfigures (a) \& (c) show the counts for intent and dialog action. Subfigure (d) show boxplots for each of the three sentiments.}
    \label{fig:stats2}
\end{figure}

The Gridspace platform records each side of the conversation separately, and we
release the audio in speaker-separated files encoded as 8kHz per the original
telephony data. We transcribed the utterances via crowd workers with basic
speech transcription training again using the Gridspace Mixer platform. Workers
are not instructed to carefully transcribe word fragments or non-speech noises.
Leveraging crowd workers and transcribing without precise fragments and
non-speech tags has been shown to be a viable approach for training speech
recognition systems \cite{novotney2010cheap}.
% We also provide estimated alignments using the Gridspace Speech Recognition API, although we did not evaluate the accuracy of these automatic alignments.
In addition to human transcriptions of each conversation, we include four additional labels:\vspace{0.8em}

\noindent\textbf{Intent}. Each conversation has a single intent representing the customer's goal in the conversation. An intent can be one of eight categories: \textit{order checks, check balance, replace card, reset password, get branch hours, pay bill, schedule appointment, transfer money}. Fig.~\ref{fig:stats2}a shows the distribution of intents to be roughly balanced. Conversation intents are derived automatically from the tasks assigned to callers during collection.\vspace{0.8em}

\noindent\textbf{Emotional Valence}. Utterances are automatically labeled with three sentiment categories, \textit{negative, neutral, and positive}. There is a probability estimate label for each category, generated by the Gridspace Speech API that is trained on a large corpus of proprietary data from multiple domains. Fig.~\ref{fig:stats2}b shows the distribution of each sentiments across utterances.\vspace{0.8em}

\noindent\textbf{Speaker ID}. Utterances have a unique ID out of 59 speakers. The number of utterances per speaker  are imbalanced, with most speakers responsible for less than 50 utterances.\vspace{0.8em}
% Fig.~\ref{fig:stats2}c shows the distribution of utterances by speaker.\vspace{0.8em}

\noindent\textbf{Dialog Action}. Every utterance is accompanies by one or more labels representing a "conversational mode". The distribution over actions is imbalanced with ``greeting'' being the most frequent and many infrequent actions combined into the ``other'' category. The 16 possible actions are: \textit{``yes'' response, greeting, response, data confirmation, procedure explanation, data question, closing, data communication, ``bear with me'' response, acknowledgement, data response, filler disfluency, thanks, open question, problem description}, and \textit{other}.  Fig.~\ref{fig:stats2}d shows the distribution of dialog actions for utterances.

\section{Spoken Language Understanding}
\label{sec:asr}

Using the \textsc{HarperValleyBank} corpus, we evaluate three common approaches for automatic speech recognition: connectionist temporal classification or CTC \cite{graves2006connectionist}, Listen-Attend-Spell or LAS \cite{chan2016listen}, and finally, a ``multi-task'' objective combining the two previous losses \cite{kim2017joint}, MTL.
% All objectives pr/oduce an \textit{embedding} vector by encoding audio features.

In addition to optimizing the speech recognition objective, denoted $\mathcal{L}_{\textup{asr}}$, we fit four linear layers mapping the encoding of the audio signal to a prediction of the intent, dialog action, and sentiment labels.
These three auxiliary objectives are optimized jointly with the speech recognition objective:
\begin{equation*}
    \beta \mathcal{L}_{\textup{asr}} + (1-\beta) \left( \mathcal{L}_{\textup{task}} + \mathcal{L}_{\textup{action}} + \mathcal{L}_{\textup{sent}} \right)
\label{eq:obj}
\end{equation*}
where $\mathcal{L}_{\textup{task}}$, and $\mathcal{L}_{\textup{sent}}$ are cross entropy losses, whereas $\mathcal{L}_{\textup{action}}$ comprises a sum of 16 binary cross entropy losses. The scalar $\beta \in [0, 1]$ weights the recognition and auxiliary objectives.
% , of which more than one may apply to any single utterance.

% \begin{table*}[h!]
% \centering
% \begin{tabular}{r c c c c c}
% \hline
% Model & WER & Speaker ID & Intent & Dialog Action (F1) & Sentiment \\
% \hline
% CTC & 46.9 & \textbf{89.5} & \textbf{47.8} & \textbf{37.4} & \textbf{82.4} \\
% LAS & 13.3 & 88.0 & 30.8 & 26.8 & 72.1 \\
% MTL & \textbf{12.7} & 88.3 & 30.7 & 27.9 & 72.4 \\
% \hline
% \end{tabular}
% \caption{Performance on the \textsc{HarperValleyBank} test set. Word error rate (WER) of speech recognition systems along with accuracy on auxiliary dialog tasks: speaker ID, caller intent, dialog action, and sentiment predition. For dialog action, we report F1 scores.
% Examples are split by utterance randomly.}
% \label{table:perf1}
% \end{table*}

%\begin{table}[h!]
%\centering
%\begin{tabular}{r c c c}
%\hline
%Model & Intent & Action & Emotion\\
%\hline
%CTC & $\mathbf{53.1}$ & $\mathbf{96.3}$ & $\mathbf{83.9}$ \\
%LAS & $39.1$ & $95.6$ & $78.1$ \\
%MTL & $42.1$ & $95.8$ & $80.7$ \\
%\hline
%\end{tabular}
%\caption{Performance on auxiliary tasks predicting the conversation intent, dialog action, and speaker emotion. Examples are split by utterance randomly.}
%\label{table:perf2}
%\end{table}

\begin{table}[b!]
\centering
\small
\begin{tabular}{r c c c c}
\hline
Model & CER &  Action (F1) & Sentiment & Intent \\
\hline
CTC & \textbf{14.43} & \textbf{0.3864} & \textbf{84.46} & \textbf{45.47} \\
LAS & 47.45 & 0.2931 & 72.09 & 34.96 \\
MTL & 38.59 & 0.3222 & 76.12 & 42.28 \\
\hline
\end{tabular}
\caption{Speech recognition and auxiliary task performance using three modern end-to-end neural approaches.}
\label{table:perf3}
\end{table}

\subsection{Training Details}

Audio is preprocessed to 128 Mel-frequency spectrogram features with a sampling rate of 8kHz, a hop length of 128, and a window size of 256.
For CTC, we encode log-Mel features using a bi-directional LSTM with two layers and 128 hidden dimensions. CTC decoding is done greedily with no language model.
In LAS, the listener network is composed of three stacked pyramid bi-directional LSTMs  with 128 hidden dimensions whereas the speller network is an uni-directional LSTM with 256 hidden dimensions and a single-headed attention layer.
For MTL, we use the same encoder as LAS but include both the speller and CTC decoder.
In MTL, we can interpret CTC as a secondary objective whose main role is to regularize the LAS encoder to respect CTC alignments:
\begin{equation*}
    \beta (\alpha\mathcal{L}_{\textup{ctc}} + (1-\alpha)\mathcal{L}_{\textup{las}}) + (1-\beta) \left( \mathcal{L}_{\textup{task}} + \mathcal{L}_{\textup{action}} + \mathcal{L}_{\textup{sent}} \right)
\label{eq:mtl}
\end{equation*}
where $\alpha$ is chosen to be 0.7 by grid search.
The LAS and MTL objectives are optimized with teacher forcing with probability 0.5.
We train each model for 200 epochs with Adam \cite{kingma2014adam} using a learning rate of 1e-3, batch size 128, and gradient clipping.

\subsection{Results and Analysis}
Table~\ref{table:perf3} shows the performance of CTC, LAS, and MTL on a speaker split test set. We report CER to measure transcription quality, accuracy for sentiment and intent prediction, and F1 for dialog action prediction (due to class imbalance). The highest performing model in each metric is bolded.

\begin{figure}[h!]
    \centering
    \begin{subfigure}[b]{0.32\linewidth}
        \centering
        \includegraphics[width=\textwidth]{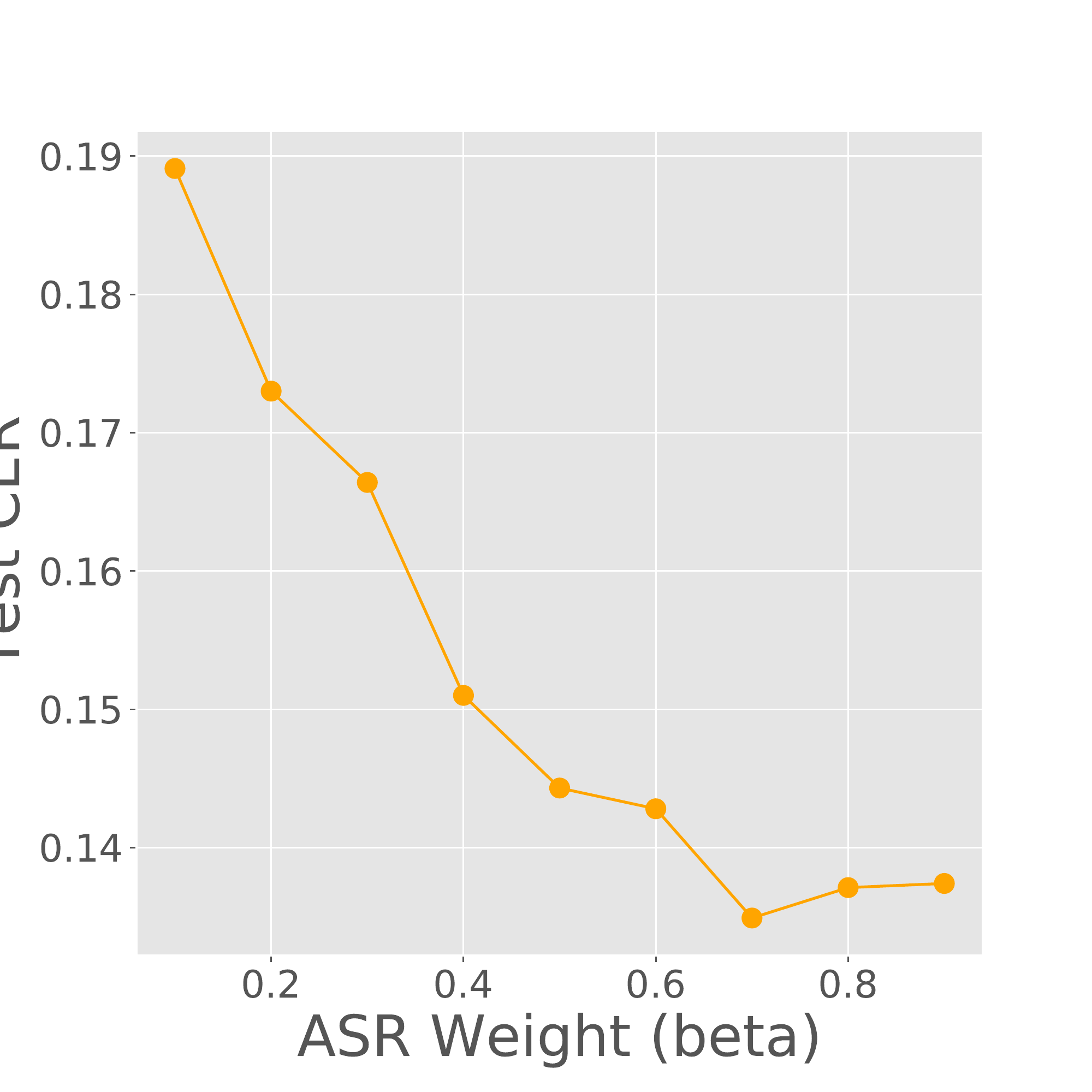}
        \caption{CTC CER}
    \end{subfigure}
    \begin{subfigure}[b]{0.32\linewidth}
        \centering
        \includegraphics[width=\textwidth]{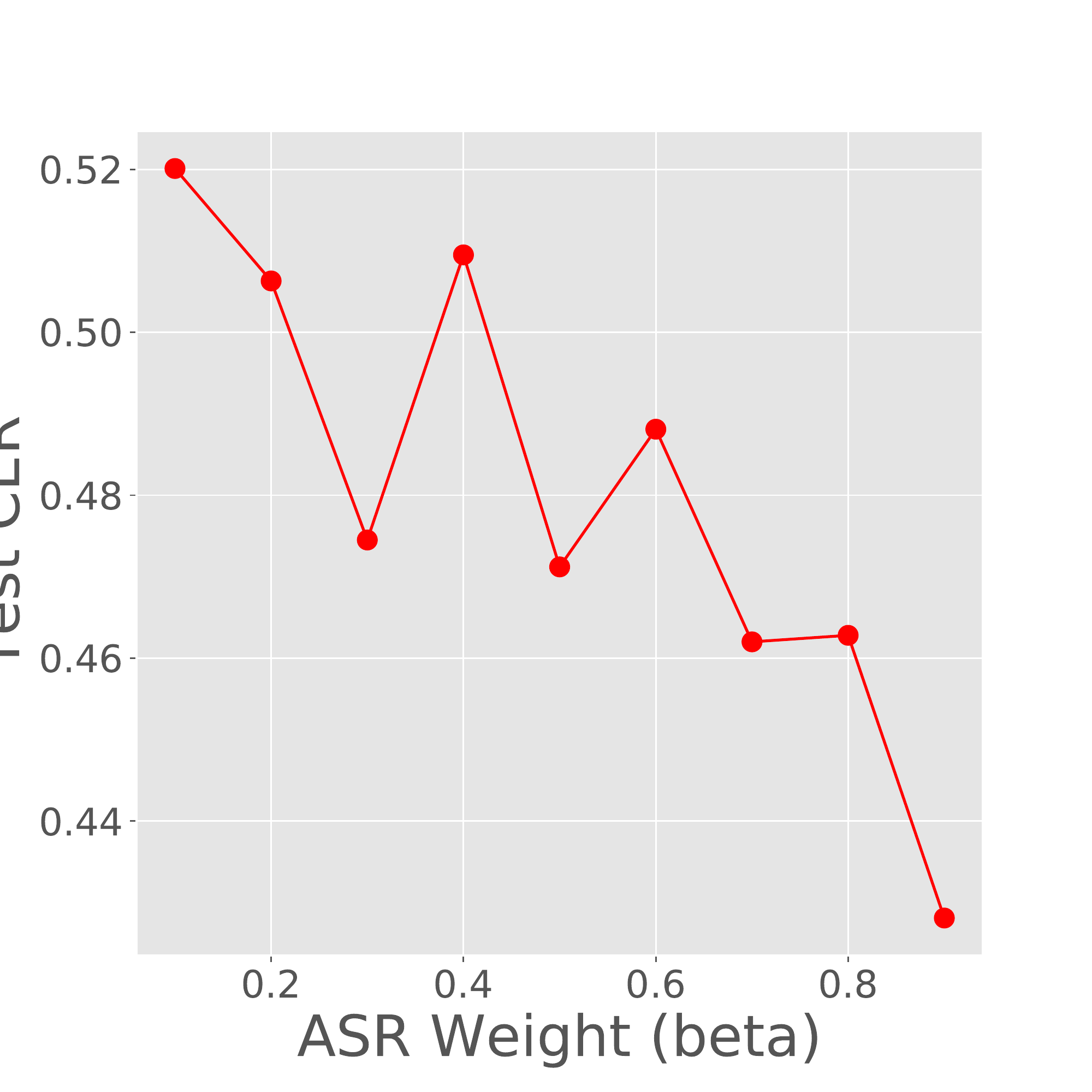}
        \caption{LAS CER}
    \end{subfigure}
    \begin{subfigure}[b]{0.32\linewidth}
        \centering
        \includegraphics[width=\textwidth]{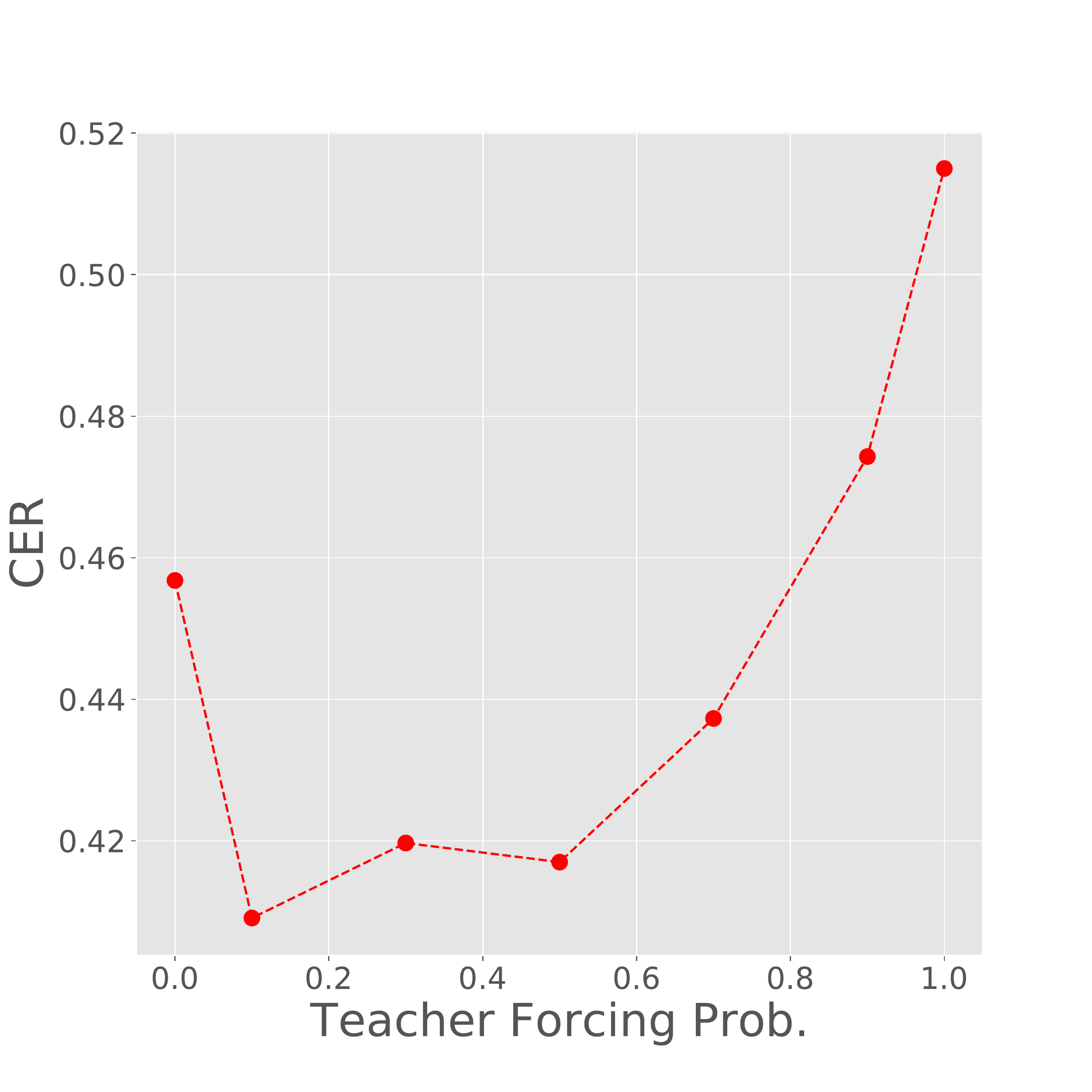}
        \caption{Teacher Forcing}
    \end{subfigure}
    \caption{Subfigure (a) and (b) show the effect of increased weight on the test CER error for CTC and LAS models. Subfigure (c) shows the effect of teacher forcing in training LAS. }
    \label{fig:stats3}
\end{figure}

Interestingly, we find that CTC outperforms LAS and MTL. We attribute this to the small size of the \textsc{HarperValleyBank} corpus that encourages expressive autoregressive models to heavily overfit. In this case, the structure of the CTC loss is beneficial compared with a model that learns attention. We see further evidence of this by MTL outperforming LAS, where CTC acts as a regularizer. LAS can improve a bit by tuning the probability of teacher forcing (see Fig~\ref{fig:stats3}).

% As consistent with prior literature, we find LAS to outperform CTC in terms of word error rate (WER) since the latter is forced to assumed conditional independence among input characters. However, we surprisingly find CTC to outperform LAS in all four auxiliary tasks, suggesting that the learned alignments enforced by CTC may be useful for speaker ID and intent prediction, among others. Further, we find MTL to outperform both LAS and CTC in word error rate, reproducing the findings from \cite{kim2017joint}, with equivalent performance as LAS on the auxiliary tasks.
% Performance for intent prediction is much lower than for the other three tasks.
% Recall that intent is the overall goal of the call (e.g. ordering a new checkbook) whereas dialog action is a function of a single utterance (e.g. asking a question). As we are fitting utterance models, predicting intent is a difficult problem.
% In contrast, Table~\ref{table:perf3} shows results on a speaker-disjoint split of the data with speaker ID results omitted by necessity. Interestingly, we no longer see CTC outperforming the end-to-end neural approaches on auxiliary tasks, suggesting that CTC has lacks generalization to unobserved speakers. Once again, we find LAS and MTL to greatly outperform CTC in WER.

\section{Unsupervised Speech Representations}
\label{sec:transfer}
Unsupervised representation learning seeks to derive useful representations of speech waveforms without any human annotations. The learned representations are reused for downstream tasks, such as predicting caller intent or dialog action.
As \textsc{HarperValleyBank} is a small dataset, it is a suitable candidate to measure the effectiveness of speech representations. In our experiments, we pretrain representations on the 100 hour split or 960 hour split of LibriSpeech \cite{panayotov2015librispeech}.

Ou baselines use recent ideas from contrastive learning \cite{wu2018unsupervised,zhuang2019local,bachman2019learning,misra2020self,he2020momentum,chen2020simple,wu2020mutual,joshi2020spanbert} where representations are learned by discriminating between specific instances in a dataset.
In particular, we adapt four algorithms from computer vision to speech: Instance Discrimination or IR \cite{wu2018unsupervised}, Local Aggregation or LA \cite{zhuang2019local}, Momentum Contrast or MoCo \cite{he2020momentum}, and SimCLR \cite{chen2020simple}.
One of our contributions is to establish comparable baselines for audio representation learning.
As a close relative, we also evaluate representations learned using Wav2Vec-1.0 \cite{schneider2019wav2vec} and Wav2Vec-2.0 \cite{baevski2020wav2vec}.

\begin{table}[h!]
\centering
\small
\begin{tabular}{r c c c c}
\hline
Model & Spk. & Intent & Action & Sent.\\
\hline
Wav2Vec 1.0 (960hr) & 18.2 & 17.1 & 0.0 & 53.7 \\
Wav2Vec 2.0 (100hr) & 22.3 & 19.7 & 0.0 & 54.3 \\
Wav2Vec 2.0 (960hr) & 27.3 & 20.5 & 0.0 & 55.5 \\
% VQ-Wav2Vec (gumbel) & & & & \\
% VQ-Wav2Vec (kmeans) & & & & \\
\hline
IR (100hr) & 99.5 & 99.1 & 0.0 & 51.3 \\
LA (100hr) & 99.5 & 98.8 & 0.0 & 50.5 \\
MoCo (100hr) & 99.6 & 98.9 & 0.0 & 53.2\\
SimCLR (100hr) & 99.8 & 99.3 & 0.0 & 53.9 \\
IR$^*$ (100hr) & 99.5 & 84.5 & 0.0 & 51.4 \\
LA$^*$ (100hr) & 97.5 & 75.8 & 1.4 & 55.1 \\
MoCo$^*$ (100hr) & 99.1 & 82.6 & 0.0 & 54.0 \\
SimCLR$^*$ (100hr) & 99.2 & 81.4 & 0.0 & 54.6 \\
\hline
IR (960hr) & 99.9 & 99.9 & 17.4 & 66.3 \\
LA (960hr) & 99.9 & 99.9 & 18.4 & 64.6 \\
MoCo (960hr) & 99.9 & 99.9 & 17.3 & 65.5 \\
SimCLR (960hr) & 99.9 & 99.9 & 17.4 & 65.9 \\
IR$^*$ (960hr) & 99.9 & 86.7 & 17.6 & 64.8 \\
LA$^*$ (960hr) & 99.9 & 79.8 & 18.0 & 64.6 \\
MoCo$^*$ (960hr) & 99.5 & 86.1 & 16.2 & 64.3 \\
SimCLR$^*$ (960hr) & 98.6 & 82.6 & 16.1 & 65.6 \\
\hline
\end{tabular}
\caption{Performance on speaker identity, intent, dialog action, and emotional valence. We report F1 score for performance on predicting dialog action. The superscript ($^*$) represents using spectral augmentations rather than wavform augmentations.}
\label{table:perf4}
\end{table}

\subsection{Training Details}
% We again ignore speakers with less than 10 utterances in the dataset.
% For Wav2Vec based models, we use the official implementation and pretrained weights available on FairSeq \cite{ott2019fairseq}.
Given a waveform, we truncate to 150k frames, and compute the log-Mel spectrogram. Spectrograms are z-scored using training statistics, which we found to be important for generalization. By default, we  augment waveforms by selecting contiguous crops with a minimum and maximum ratio of 0.08 to 1.0, along with Gaussian noise with a scale of 1.0. We separately explore first computing the spectrogram, then applying a time and frequency mask as augmentation, denoted by the ($^*$) superscript in Table~\ref{table:perf4}. Regardless, we fit a ResNet50 \cite{he2016deep} to map the spectrogram to a 2048 dimensional embedding.
% Wav2Vec-1.0 and Wav2Vec-2.0 use customized architectures that amount to around 1.5 times the number of parameters as ResNet50.

After representation learning, we measure the quality of an embedding by linear evaluation \cite{wu2018unsupervised,zhuang2019local,he2020momentum,chen2020simple}.
% We use the features after the last convolutional layer and prior to the global pooling, resulting in a 2048x4x4 dimensional vector and fit a logistic regression model mapping this vector to a probability for each class in the transfer task. A separate  regression is fit for speaker ID, intent, dialog action, and emotional valence.
% For the Wav2Vec family, we use the encoded embedding, averaged over timesteps, with dimension 512 and 1024 for the 1.0 and 2.0 models, respectively.
The \textsc{HarperValleyBank} corpus is split into train (80\%) and test (20\%) sets \textit{by class} to ensure both sets have instances of each class.
Thus, each transfer task has its own train test split. Refer to the public repository for training  hyperparameters.
% For dialog actions, we ignore all actions with less than 200 appearances.
% The same data augmentations used in pretraining are used in transfer but not in evaluation.

% In optimization, we use SGD with batch size 256, learning rate 0.03, momentum 0.9, weight decay 1e-4 for 200 epochs. In transfer, we use SGD with batch size 256, learning rate 0.01, momentum 0.9, weight decay 1e-5 for 100 epochs. In the contrastive objectives, we use a temperature $\tau{=}0.07$. For IR and LA, we use 4096 negative samples and set the memory bank update parameter to 0.5. For LA, we fit KMeans with $k{=}5000$ ten times with different random seeds and take the union of all of 10 clusters with the current input as a member. For MoCo, we use a queue of size 66536 and a momentum update of 0.99.
% We leave more careful hyperparameter search to future work.

\subsection{Results and Analysis}
Table~\ref{table:perf4} compares the different unsupervised models: IR, LA, MoCo, and SimCLR surpass supervised methods (e.g. CTC, LAS, and MTL) in intent prediction,  falling short in dialog action and sentiment prediction. In particular, dialog action prediction is a surprisingly difficult task for our unsupervised representations. Whereas supervised methods achieve upwards of 0.3 F1, the best models in Table~\ref{table:perf4} achieve only half the score, despite seeing 960 hours of speech. That being said, adding 860 hours of speech improved the representations, as shown by overall better performance. Finally, compared to Wav2Vec, the contrastive objectives find large gains of up to 70\%.

% In summary, we proposed three baseline algorithms for representation learning in audio with the caller intents from \textsc{HarperValleyBank} as measures of  the usefulness of a representation, which future research can build upon.

\section{Conclusion}

We introduced \textsc{HarperValleyBank}, a new speech corpus of transcribed
conversations between employees and customers in a bank transcaction. The
corpus includes additional labels, including speaker identity, caller intent,
dialog actions, and sentiment. In our experiments, we established baseline
models that showed this corpus to be an interesting challenge for future
algorithms, and a useful educational tool for modern deep learning approaches to spoken dialog.
Our experiments analyzed utterances independently, future work can explore using the \textsc{HarperValleyBank}
corpus in conversation modelling and its related downstream optimization.

\section{Acknowledgments}
We thank the Gridspace team for their support in data collection and analysis.
Experiments used PyTorch, PyTorch Lightning, and Weights \&
Biases \cite{biewald2020experiment}.
We thank Samuel Kwong and Dan Jurafsky for helpful discussion and feedback.

\bibliographystyle{IEEEtran}

\bibliography{mybib}

\end{document}